\documentclass{article}
\usepackage{spconf,amsmath,graphicx,hyperref}
\usepackage{amssymb} 
\usepackage{tabularx} 
\usepackage{multirow}
\usepackage{makecell}
\usepackage{array}

\title{Enhancing Few-Shot Time Series Forecasting with LLM-Guided Diffusion}
\name{Haonan Shi$^{1 \dagger}$ , Dehua Shuai$^{2 \dagger}$ , Liming Wang$^{1}$ ,  Xiyang Liu$^{1 \ddagger}$   , Long Tian$^{1 \ddagger}$ 
\thanks{$^{\dagger}$ These authors contributed equally.}
\thanks{$^{\ddagger}$ Corresponding authors: \{xyliu, tianlong\}@xidian.edu.cn}
}

\address{
$^1$ School of Computer Science and Technology, Xidian University, Xi'an Shi, China \\[0.5em]
$^2$ Hangzhou Institute of Technology, Xidian University, Hang Zhou Shi, China \\[0.5em]
}

\begin{document}
%
\maketitle
\begin{abstract}
Time series forecasting in specialized domains is often constrained by limited data availability, where conventional models typically require large-scale datasets to effectively capture underlying temporal dynamics. To tackle this few-shot challenge, we propose LTSM-DIFF (Large-scale Temporal Sequential Memory with Diffusion), a novel learning framework that integrates the expressive power of large language models with the generative capability of diffusion models. Specifically, the LTSM module is fine-tuned and employed as a temporal memory mechanism, extracting rich sequential representations even under data-scarce conditions. These representations are then utilized as conditional guidance for a joint probability diffusion process, enabling refined modeling of complex temporal patterns. This design allows knowledge transfer from the language domain to time series tasks, substantially enhancing both generalization and robustness. Extensive experiments across diverse benchmarks demonstrate that LTSM-DIFF consistently achieves state-of-the-art performance in data-rich scenarios, while also delivering significant improvements in few-shot forecasting. Our work establishes a new paradigm for time series analysis under data scarcity.
\end{abstract}
\section{Introduction}
\label{sec:intro}

Time series forecasting is a cornerstone of data analysis, with critical applications spanning finance, meteorology, and healthcare~\cite{wang2024deep}. While deep learning models have achieved remarkable success, their performance is often contingent on the availability of extensive historical data~\cite{10887797,zhang2025multi}. In many specialized domains, however, data is scarce, posing a significant challenge known as few-shot time series forecasting. Traditional models, trained from scratch, struggle in these data-limited environments as they fail to capture complex temporal dynamics and generalize effectively, leading to suboptimal predictive accuracy~\cite{chang_llm4ts_2025}.

To overcome the limitations of data scarcity, recent research has explored the adaptation of pre-trained Large Language Models (LLMs) for time series analysis~\cite{chang_llm4ts_2025,jin_time-llm_2024, gruver_large_2024}. By leveraging their profound capabilities in pattern recognition and sequential data processing, LLMs have demonstrated promise in zero-shot and few-shot settings~\cite{gruver_large_2024}. These models involve reprogramming time series as sequences of tokens or using textual prompts to guide the forecasting process~\cite{tan_are_nodate}. However, a fundamental modality gap exists between continuous numerical time series and the discrete textual domain for which LLMs were originally trained. This discrepancy can lead to challenges in accurately capturing fine-grained temporal patterns and may limit the precision of the forecast~\cite{liu_calf_nodate}.

Concurrently, diffusion models have emerged as a powerful class of generative models, excelling at learning complex data distributions and generating high-fidelity samples~\cite{yang_diffusion_2023}. Their application to time series has yielded state-of-the-art results in probabilistic forecasting and imputation by modeling the data distribution through a gradual denoising process~\cite{tashiro_csdi_2021, shen_non-autoregressive_2023}. These models are inherently probabilistic, providing valuable uncertainty estimates alongside predictions. Nevertheless, their effectiveness relies on a robust conditioning mechanism to guide the generation process based on historical context. In few-shot scenarios, learning this complex conditional distribution from limited data remains a formidable challenge, affecting their generative power~\cite{kollovieh_predict_2023}.

In this paper, we propose \textbf{LTSM-DIFF} (Large-scale Temporal Sequential Memory with Diffusion), a novel framework that integrates a LLM as a \textbf{temporal memory module} with a conditional diffusion model for generative time series forecasting. Using the LLM to extract rich sequential representations, our approach effectively captures long-term dependencies and complex temporal patterns, even in data-scarce scenarios. These representations guide a joint probability diffusion process, refining the predictive distribution and enabling cross-domain knowledge transfer from language to numerical time series. Extensive experiments on diverse benchmarks demonstrate that LTSM-DIFF achieves state-of-the-art performance in both data-rich and few-shot settings, establishing a new paradigm for forecasting with limited data.

\section{Related Work}
\label{sec:Related}

\subsection{Time Series Forecasting with LLMs} 
Large Language Models (LLMs) have recently demonstrated strong potential for time series forecasting due to their ability to capture sequential dependencies. A central challenge, however, lies in addressing the modality gap between discrete natural language tokens and continuous numerical time series data. Early studies attempted to mitigate this gap by reprogramming time series into text-like sequences \cite{jin_time-llm_2024} or directly mapping numerical values into textual tokens \cite{gruver_large_2024}. Building upon this, more advanced frameworks such as LLM4TS \cite{chang_llm4ts_2025} and CALF \cite{liu_calf_nodate} introduced specialized fine-tuning schemes and cross-modal alignment techniques to adapt pre-trained LLMs to temporal dynamics. Additional efforts have explored enhancing forecasting accuracy by integrating external sources, for example through retrieval-augmented generation (RAG) \cite{yang_timerag_2024} or multi-modal data fusion with text and vision \cite{zhong_time-vlm_2025}. While these methods highlight the versatility of LLMs, they often remain limited by their reliance on token-level reprogramming or external augmentation. 
In contrast, our LTSM-DIFF framework employs a fine-tuned LLM as a dedicated temporal memory module, which provides structured conditional guidance to a downstream diffusion process, with the goal of improving few-shot forecasting performance.

\subsection{Time Series Forecasting with Diffusion Models} 
Diffusion models have recently emerged as a promising paradigm for probabilistic time series forecasting, due to their capacity to model complex distributions and quantify predictive uncertainty. Most existing approaches adopt conditional diffusion frameworks, where historical observations are used as conditioning signals for generating future trajectories \cite{shen_non-autoregressive_2023}. To enhance temporal modeling, researchers have combined diffusion mechanisms with sequence models such as Transformers \cite{li_transformer-modulated_2024} and structured state-space models \cite{alcaraz_diffusion-based_nodate}. Other works improve the generative process through enhanced guidance strategies, including self-guidance \cite{kollovieh_predict_2023}, multi-granularity learning \cite{fan_mg-tsd_2024}, and retrieval augmentation \cite{liu_retrieval-augmented_2024}. Despite these advances, conditioning remains primarily derived from the raw time series, which can be unreliable in data-scarce scenarios. 
LTSM-DIFF addresses this limitation by introducing a cross-modal conditioning mechanism, where high-level semantic representations from an LLM guide the diffusion process. This approach is designed to enable more effective forecasting under few-shot conditions.

\begin{figure*}[t]
    \centering
    \includegraphics[scale=0.7,trim=4.5cm 3.05cm 4.5cm 3.05cm,clip]{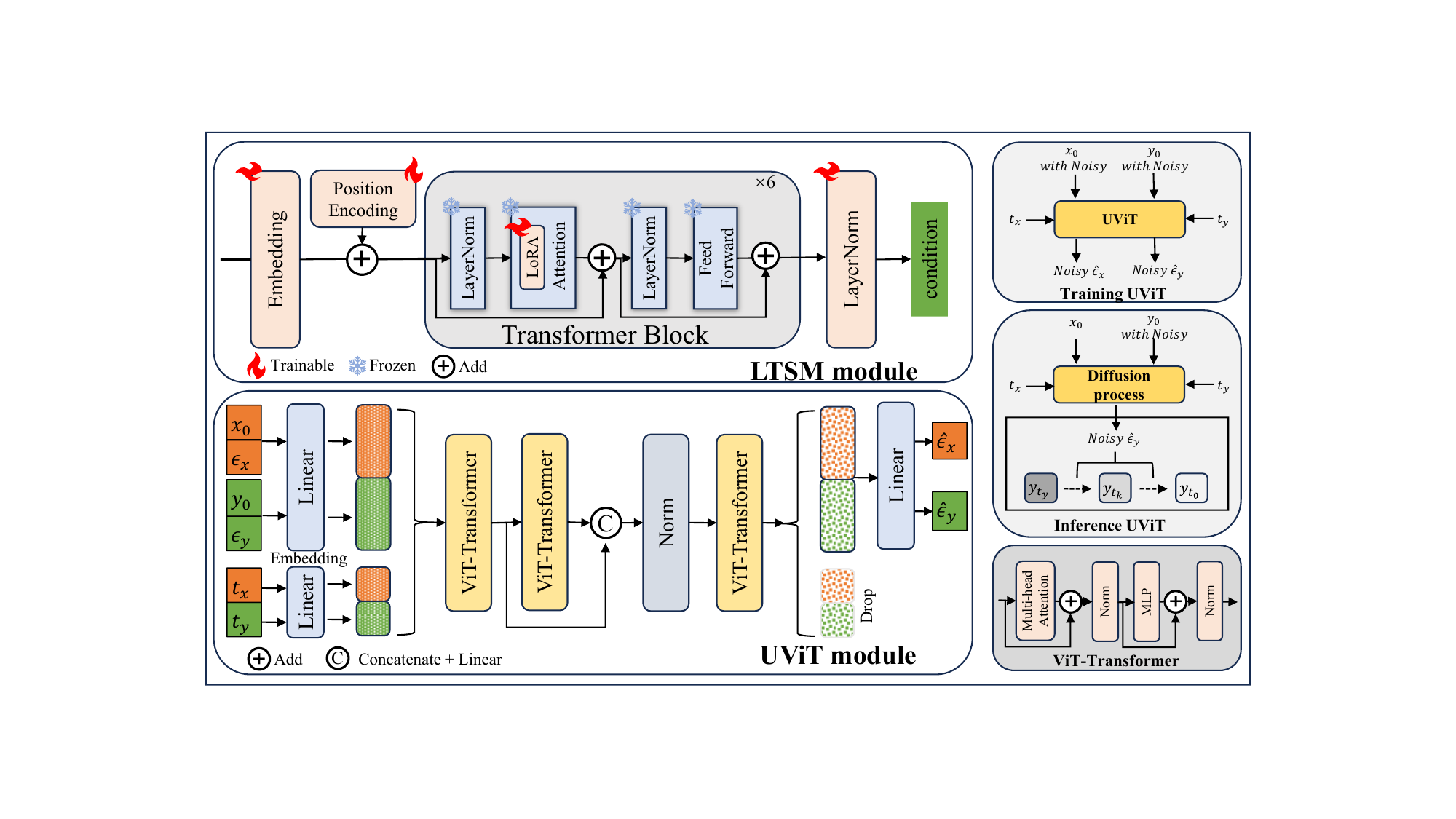}
    \caption{Overall framework of our proposed method. 
    The encoder based on a foundation model extracts temporal representations, 
    while the UViT-based diffusion module performs conditional forecasting. }
    \label{fig:method}
\end{figure*}
\vspace{-1em}

\section{Method}
\label{sec:Method}

In this section, we present our proposed framework for multivariate time series forecasting. 
As illustrated in Fig.~\ref{fig:method}, the overall architecture consists of two key components: 
(1) a foundation-model-based encoder for extracting temporal representations, and 
(2) a UViT-based diffusion module for conditional forecasting. 
The following subsections describe the details of each component and the joint training strategy.

\subsection{Problem Definition}

In multivariate time series forecasting, given a historical data:
\begin{equation}
\mathbf{X}_{1:T} = \{\mathbf{x}_1,\ldots,\mathbf{x}_T\}, \quad \mathbf{x}_t \in \mathbb{R}^d,
\end{equation}
the goal is to predict the next $H$ steps:
\begin{equation}
\mathbf{Y}_{T+1:T+H} = \{\mathbf{y}_{T+1},\ldots,\mathbf{y}_{T+H}\}, \quad \mathbf{y}_t \in \mathbb{R}^d,
\end{equation}
which amounts to modeling the conditional distribution:
\begin{equation}
p(\mathbf{Y}_{T+1:T+H} \mid \mathbf{X}_{1:T}).
\end{equation}

\subsection{Temporal Representation with a Foundation Model}
To obtain a powerful temporal prior, we employ an encoder based on a foundation model. 
The input sequence $\mathbf{X}_{1:T}$ first passes through an Embedding module, which linearly projects the $d$-dimensional features into an $m$-dimensional embedding space and then uses a Transformer encoder layer to capture local dependencies.

These embeddings are subsequently fed into the first six Transformer blocks of a pre-trained GPT-2 model. To efficiently adapt GPT-2 to the time-series domain, we adopt Low-Rank Adaptation (LoRA): small, trainable low-rank matrices are injected into the query, key, and value projection layers  of the self-attention mechanism, while the original GPT-2 weights remain frozen. This design keeps the number of trainable parameters small while effectively adapting the model to time-series data. The resulting sequence of hidden states serves as a rich temporal representation:
\begin{equation}
\mathbf{x_0} = f_{\text{LLM-Embedding}}(\mathbf{X}_{1:T}) \in \mathbb{R}^{T \times m}.
\end{equation}
This entire encoder module is pre-trained on an autoregressive forecasting task using a mean squared error (MSE) loss to establish its predictive capability:
\begin{equation}
\mathcal{L}_{\text{LLM}} = \frac{1}{H}\sum_{t=T+1}^{T+H}\|\hat{\mathbf{y}}_t - \mathbf{y}_t\|_2^2.
\end{equation}

\subsection{Conditional Forecasting with a Diffusion Model}
Inspired by UniDiffuser~\cite{bao_one_2023}, we design a conditional diffusion forecaster based on UViT, a U-Net variant that replaces convolutional layers with Transformer blocks. Unlike standard conditional diffusion, we apply the forward noising process to both the conditions and the future targets. This strategy enables the model to learn their joint distribution more flexibly, while also allowing it to handle noisy or missing conditions and to support both conditional and unconditional generation.
The forward processes are defined as:
\begin{align}
q(\mathbf{x}_{t_x} \mid \mathbf{x}_0) &= \sqrt{\bar{\alpha}_{t_x}}\mathbf{x}_0 + \sqrt{1-\bar{\alpha}_{t_x}}\,\epsilon_x, \\
q(\mathbf{y}_{t_y} \mid \mathbf{y}_0) &= \sqrt{\bar{\alpha}_{t_y}}\mathbf{y}_0 + \sqrt{1-\bar{\alpha}_{t_y}}\,\epsilon_y,
\end{align}

where $\bar{\alpha}$ is the noise scheduling parameter, $\epsilon_x,\epsilon_y \sim \mathcal{N}(0,\mathbf{I})$ and $t_x,t_y$ are uniformly sampled timesteps.

The UViT network is trained to predict the injected noise $[\epsilon_x, \epsilon_y]$ from their noisy versions $[\mathbf{x}_{t_x}, \mathbf{y}_{t_y}]$: 

\begin{equation}
[\hat{\epsilon}_x,\hat{\epsilon}_y] = \text{UViT}(\mathbf{x}_{t_x}, \mathbf{y}_{t_y}, t_x, t_y ),
\end{equation}

with the training objective being the noise matching loss:
\begin{equation}
\mathcal{L}_{\text{diff}} = \mathbb{E}_{\mathbf{x}_0,\mathbf{y}_0,\epsilon_x,\epsilon_y,t_x,t_y}\left[\| \epsilon_x-\hat{\epsilon}_x \|_2^2 + \| \epsilon_y-\hat{\epsilon}_y \|_2^2 \right].
\end{equation}

By setting the condition's timestep $t_x=0$, the model performs standard conditional forecasting. Other combinations of $(t_x,t_y)$ encourage it to learn the marginal and joint distributions.

During inference, the model generates the target $\mathbf{y}_0$ given a condition $\mathbf{x}_0$ through an iterative denoising process. Starting from pure noise $\mathbf{y}_T \sim \mathcal{N}(0, \mathbf{I})$, at each reverse step $t$ from $T$ down to $1$, the model predicts the noise conditioned on the clean input:
$
\hat{\epsilon}_y = \text{UViT}(\mathbf{x}_0, \mathbf{y}_t, 0, t )
$,
This predicted noise is then used to obtain the less noisy sample $\mathbf{y}_{t-1}$ by applying the standard update formulas from established samplers like DDPM~\cite{ho2020denoising} or DDIM~\cite{song2020denoising}. Repeating this process until $t=0$ yields the final clean sample $\mathbf{y}_0$.

\begin{table*}[ht]
\centering
\caption{Full-data forecasting performance comparison across datasets. Best results are highlighted in \textbf{bold}.}
\label{tab:full}
\resizebox{\textwidth}{!}{%
\normalsize
\begin{tabular}{|l|c c|c c|c c|c c|c c|c c|c c|c c|c c|c c|}
\hline
\textbf{Models} & \multicolumn{2}{c|}{\textbf{LTSM-DIFF}} & \multicolumn{2}{c|}{TimeLLM} & \multicolumn{2}{c|}{GPT4TS} & \multicolumn{2}{c|}{CALF} & \multicolumn{2}{c|}{PatchTST} & \multicolumn{2}{c|}{iTransformer} & \multicolumn{2}{c|}{Crossformer}  & \multicolumn{2}{c|}{TMDM} & \multicolumn{2}{c|}{TimeGrad} & \multicolumn{2}{c|}{TimeDiff} \\
\textbf{Dataset}
& MSE & MAE & MSE & MAE & MSE & MAE & MSE & MAE & MSE & MAE & MSE & MAE & MSE & MAE & MSE & MAE & MSE & MAE & MSE & MAE \\
\hline
Electricity& \textbf{0.172} & \textbf{0.264}& 0.223 & 0.309 & 0.205 & 0.290 & 0.176 & 0.266 & 0.216 & 0.304 & 0.178 & 0.270 & 0.244 & 0.334  & 0.190 & 0.270 & 0.690 & 0.740 & 0.270 & 0.320 \\
\hline
Traffic& 0.433& \textbf{0.279}& 0.541 & 0.358 & 0.488 & 0.317 & 0.439 & 0.282 & 0.555 & 0.361 & \textbf{0.428}& 0.282 & 0.550 & 0.304  & 0.600 & 0.350 & 0.960 & 0.810 & 0.680 & 0.470 \\
\hline
ETTh1& \textbf{0.441} & \textbf{0.435}& 0.460& 0.449& 0.447 & 0.436 & 0.445& 0.436 & 0.450 & 0.441 & 0.455& 0.448 & 0.620 & 0.572  & - & - & - & - & - & - \\
\hline
ETTh2& \textbf{0.365}& \textbf{0.393}& 0.389 & 0.408 & 0.381 & 0.408 & 0.377& 0.399 & 0.366& 0.394& 0.381 & 0.405 & 0.942 & 0.684  & - & - & - & - & - & - \\
\hline
ETTm1& 0.403& 0.402& 0.410 & 0.409 & 0.389& 0.397& 0.409 & 0.404 & \textbf{0.381}& \textbf{0.395}& 0.407 & 0.411 & 0.502 & 0.502  & - & - & - & - & - & - \\
\hline
ETTm2 & \textbf{0.283}& \textbf{0.327}& 0.296 & 0.340 & 0.285& 0.331& 0.285 & 0.328& 0.285& 0.327 & 0.291 & 0.335 & 1.216 & 0.707 & 0.330& 0.400& 1.360 & 0.740 & 0.410 & 0.420 \\
\hline
Weather & \textbf{0.248} & \textbf{0.273} & 0.274 & 0.290 & 0.264 & 0.284 & 0.252 & 0.276 & 0.258 & 0.280 & 0.257 & 0.279 & 0.259 & 0.315  & 0.280 & 0.250 & 0.900 & 0.570 & 0.360 & 0.370 \\
\hline
\end{tabular}
}
\vspace{-1em}
\end{table*}

\subsection{Joint Training Objective}
The final loss combines the autoregressive and diffusion objectives:
\begin{equation}
\mathcal{L} = \mathcal{L}_{\text{LLM}} + \lambda \mathcal{L}_{\text{diff}},
\end{equation}

where $\lambda$ is a balancing hyperparameter. 

\section{Experiment}
\label{sec:Experimental}
\subsection{Experimental Setup}
All experiments are conducted on a single NVIDIA L20 GPU. Forecasting performance is measured by Mean Squared Error (MSE) and Mean Absolute Error (MAE).
We compare against representative baselines from three categories:
(1) \textbf{LLM-based models}: TimeLLM \cite{jin_time-llm_2024}, GPT4TS \cite{chang_llm4ts_2025}, and CALF \cite{liu_calf_nodate};  
(2) \textbf{Transformer-based models}: PatchTST \cite{nie2022time}, iTransformer \cite{liu2023itransformer}, Crossformer \cite{zhang2023crossformer};
(3) \textbf{Diffusion-based models}: TMDM \cite{li_transformer-modulated_2024}, TimeGrad  \cite{rasul2021autoregressive}, and TimeDiff \cite{shen_non-autoregressive_2023}.  

We follow the dataset settings of~\cite{wu2021autoformer}, evaluating on seven real-world datasets: ETT (ETTh1, ETTh2, ETTm1, ETTm2), Weather, ECL, and Traffic. Input series are truncated to $T=96$, and results are averaged over prediction horizons $H \in \{96,192,336,720\}$.

The implementation details are:
\textbf{General}: Batch Size=32, Learning Rate=0.0005, Max Epochs=20, Patience=5, Heads=4, Dropout=0.3, GPT-2 Layers=6.
\textbf{LoRA config}: r=8, alpha=32, Dropout=0.1.

\subsection{Full-data Comparison}
We first evaluate LTSM-DIFF against all baselines using the full training set. As shown in Table~\ref{tab:full}, our method consistently achieves the lowest MSE and MAE across multiple datasets, demonstrating its superior ability to capture temporal dependencies and model uncertainty effectively.

\subsection{Few-shot Transfer}
To evaluate the robustness of our model under limited supervision, we conduct few-shot transfer experiments. LTSM-DIFF is first fully pre-trained on a source dataset and then fine-tuned on 1\%, 5\%, 50\%, and 100\% of a target dataset. As shown in Table~\ref{tab:fewshot}, the results indicate that even with very small amounts of fine-tuning data (1\% or 5\%), our model maintains comparable performance to full-data fine-tuning, demonstrating the stability of LLM-guided diffusion as a temporal memory mechanism during transfer.

\begin{table}[h]
\centering
\caption{Few-shot fine-tuning performance (ours only).}
\label{tab:fewshot}
\resizebox{\columnwidth}{!}{%
\normalsize
\begin{tabular}{|c|c|c|c|c|c|}
\hline
\textbf{Transfer} & \textbf{Ratio} & \textbf{1\%} & \textbf{5\%} & \textbf{50\%} & \textbf{100\%} \\
\hline
\multirow{2}{*}{ETTm1$\to$ETTm2} 
& MSE & 0.263 & 0.263 & 0.262 & 0.262 \\
& MAE & 0.314 & 0.314 & 0.314 & 0.317 \\
\hline
\multirow{2}{*}{ETTm2$\to$ETTm1} 
& MSE & 0.679& 0.679&  0.677&  0.675\\
& MAE &  0.511&  0.511&  0.510&  0.510\\
\hline
\end{tabular}
}
\vspace{-2em}
\end{table}




\begin{table}[h]
\centering
\caption{Ablation study results with module inclusion.}
\label{tab:ablation}
\resizebox{\columnwidth}{!}{%
\normalsize
\begin{tabular}{|c|c|c|c|c|c|}
\hline
\textbf{LTSM} & \textbf{Autoformer} & \textbf{RAG} & \textbf{Diffusion} & \textbf{MSE} & \textbf{MAE} \\
\hline
\checkmark &  &  &  & 0.279 & 0.306 \\
& \checkmark &  & \checkmark & 0.357 & 0.394 \\
 &  & \checkmark & \checkmark & 0.531 & 0.434 \\
\checkmark &  &  & \checkmark & \textbf{0.248} & \textbf{0.273} \\
\hline
\end{tabular}%
}
\vspace{-1em}
\end{table}


\subsection{Ablation Study}
We conduct an ablation study to analyze the contribution of each component in our framework.  
Specifically, we compare four settings:  
(1) LLM fine-tuning only;  
(2) Autoformer + diffusion;  
(3) Few-shot retrieval (5\% storage RAG)\cite{liu_retrieval-augmented_2024} + diffusion;  
(4) Full model (LLM + diffusion).  

The results of the ${weather}$ dataset in Table~\ref{tab:ablation} show that each component brings incremental improvement, with the final combination achieving the best performance.

\begin{figure}[h!]
    \centering
    \includegraphics[
        width=\columnwidth,           
        trim={0cm 1.3cm 0cm 0cm},        
        clip                               
    ]{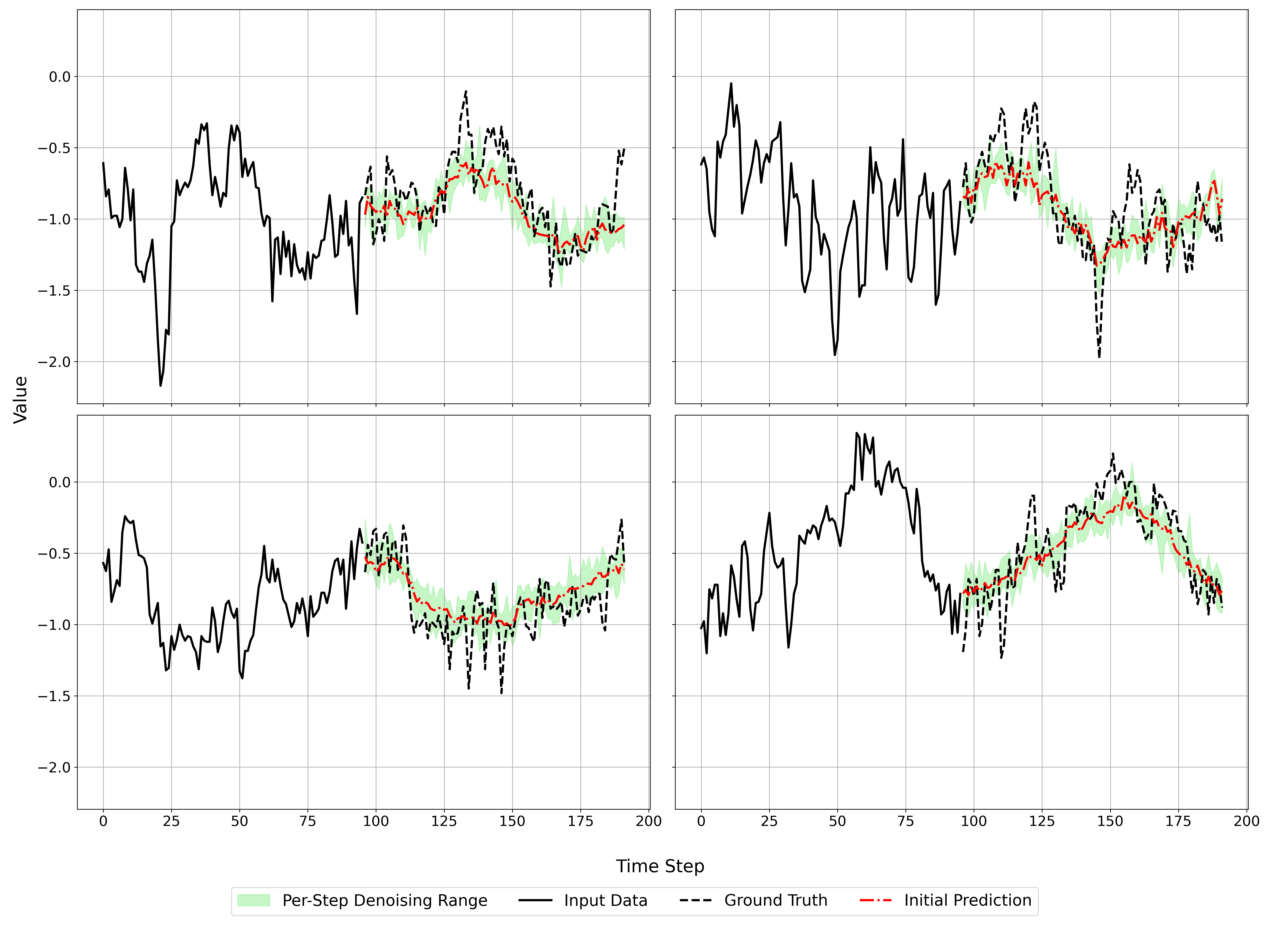} 
    \caption{
    Visualization of the denoising process on an ETTm2 sample: black lines represent historical input and ground truth, the red line is the initial LTSM prediction, and the green shaded band shows the diffusion uncertainty range.
    }
    \label{fig:denoising_process}
\end{figure}

Refer to this idea\cite{bian_multi_patch_2024}, which suggests shallow layers better capture coarse-grained temporal tokens. Our ablation study confirms that Layer-6 offers the optimal trade-off between error (MSE/MAE) and efficiency:
\begin{table}[h]
\centering
\caption{Ablation on GPT-2 layers}
\begin{tabular}{cccccc}
\hline
Layer & 1 & 3 & 6 & 9 & 12 \\
\hline
MSE & 0.381 & 0.378 & \textbf{0.377} & 0.381 & 0.379 \\
MAE & 0.398 & 0.395 & \textbf{0.394} & 0.399 & 0.397 \\
\hline
\end{tabular}
\end{table}

\subsection{Visualization of the Denoising Process}
\label{sec:vis_denoising}

As illustrated in Figure~\ref{fig:denoising_process}, the LTSM module provides an effective guiding representation that captures the primary trend. The diffusion model then refines this initial prediction by exploring a bounded uncertainty space to yield a more accurate forecast.

\section{Conclusion}

We introduce LTSM-DIFF, a novel framework for time series forecasting that synergistically combines a fine-tuned Large Language Model (LLM) as a temporal memory module with a joint probability diffusion model. Our approach achieves superior performance in data-rich scenarios and maintains high accuracy in few-shot settings, as validated by comprehensive experiments and ablation studies. This work establishes a promising paradigm for transferring knowledge from pre-trained language models to enhance time series forecasting.





\small
\bibliographystyle{IEEEbib}
\bibliography{strings,refs}

\end{document}